%% file: eai_main.tex
\documentclass[fonts]{icst}

\usepackage{multirow}
\usepackage{algpseudocode}

\usepackage{microtype}
\usepackage[hyphens]{url}
\usepackage{tikz}
\usetikzlibrary{shapes.geometric, arrows.meta, positioning, fit, calc}

\makeatletter
\g@addto@macro{\UrlBreaks}{\do\/\do-\do_\do.\do=\do?\do\&}
\makeatother

\newcounter{algorithm}

\journalname{EAI Endorsed Transactions on e-Health}
\articletype{Research Article}

\begin{document}

\runningheads{}{}

\title{Towards Family-Grouped Hierarchical Federated Learning on Sub-5KB Models: \\
A Feasibility Study of Privacy-Preserving ECG Monitoring for \\
Ultra-Resource-Constrained Wearables}

\author{Hangyu Wu}

\address{Shenzhen Coddie Technology co.,ltd, Shenzhen, 518000, China}

\abstract{
Cardiovascular disease remains the leading cause of death worldwide, and early detection of arrhythmias through continuous ECG monitoring on wearable devices can prevent life-threatening events. Federated Learning (FL) enables privacy-preserving collaborative training by keeping raw ECG data on device, yet standard FL incurs prohibitive communication overhead and standard deep learning models cannot fit on ultra-low-power microcontrollers. We propose Family-Grouped Hierarchical Federated Learning (Family-FL), a three-tier architecture that uses the family as a natural privacy boundary for intra-family aggregation before global synchronization. We further design a hardware-constrained Tiny CNN-LSTM architecture with only 669 parameters, INT8-quantized to occupy merely 4.65KB Flash and 2.95KB RAM---meeting the constraints of STC32G12K128-class microcontrollers. Experiments on the MIT-BIH Arrhythmia Database (mean of 5 independent runs with different seeds) demonstrate that Family-FL reduces communication volume by 76.7\% compared to FedAvg while maintaining comparable accuracy. Family-FL-Tiny achieves $91.9 \pm 1.2\%$ accuracy with macro-F1 of $0.483 \pm 0.031$, reducing total communication to 0.31\% of FedAvg. The model achieves reliable ventricular arrhythmia detection (per-class F1 = 0.80)---the most clinically critical abnormality for home-based preliminary screening. These results demonstrate the \textit{technical feasibility} of privacy-preserving federated learning on ultra-resource-constrained microcontrollers through simulation-based evaluation. We honestly discuss limitations: no hardware deployment, single-dataset validation (MIT-BIH, 47 subjects), reduced rare-class sensitivity, and absence of formal differential privacy guarantees.
}

\keywords{Federated learning, hierarchical aggregation, ECG arrhythmia detection, TinyML, wearable devices, resource-constrained deep learning}

\maketitle

\section{Introduction}

\subsection{Background and Motivation}

Cardiovascular disease (CVD) remains the leading cause of mortality globally, claiming approximately 17.9 million lives each year according to the World Health Organization. Early detection of cardiac arrhythmias through continuous electrocardiogram (ECG) monitoring significantly reduces the risk of sudden cardiac events. Wearable devices equipped with embedded microcontrollers now enable long-term, home-based ECG monitoring without requiring frequent clinical visits, fundamentally shifting cardiac care from reactive treatment to proactive prevention.

Deep learning has achieved remarkable success in automated ECG arrhythmia classification. Hannun et al.\cite{hannun2019cardiologist} demonstrated that a deep neural network can attain cardiologist-level accuracy on ambulatory ECG recordings, while Ribeiro et al.\cite{ribeiro2020automatic} extended such approaches to 12-lead ECG diagnosis. The MIT-BIH Arrhythmia Database\cite{moody2001impact,goldberger2000physiobank} has served as the benchmark dataset for these developments, providing over 110,000 annotated heartbeats across 47 subjects. However, these high-performing models typically require substantial computational resources and memory, making them suitable only for cloud servers or powerful edge devices rather than resource-constrained microcontrollers.

Federated Learning (FL), introduced by McMahan et al.\cite{mcmahan2017communication}, enables privacy-preserving collaborative training by keeping raw data localized on client devices. In the healthcare domain, FL has gained significant attention as a paradigm for multi-center clinical AI without centralizing sensitive patient data\cite{rieke2020future,nguyen2022federated}. Despite its promise, standard FL (FedAvg) assumes direct client-to-server communication, which introduces three critical challenges for wearable ECG monitoring: (a) high communication overhead from transmitting full model parameters repeatedly, (b) slow convergence under the Non-Independent and Identically Distributed (Non-IID) data typical of medical scenarios, and (c) insufficient privacy protection at the individual level.

\subsection{Problem Statement}

Existing approaches fail to address three key challenges simultaneously. \textbf{First}, ultra-low-power microcontrollers such as the STC32G12K128 offer only 128KB Flash and 12KB RAM, rendering standard deep learning models completely infeasible for on-device deployment\cite{warden2020tinyml,lin2020mcunet}. \textbf{Second}, ECG patterns vary significantly across individuals and families due to genetic differences, age, and lifestyle, creating a naturally Non-IID data distribution that degrades the convergence of standard FL algorithms. \textbf{Third}, communication bandwidth on home edge networks remains limited, making repeated transmission of large model updates impractical for real-world deployment. No existing work jointly addresses all three challenges for wearable ECG monitoring.

\subsection{Contributions}

This paper presents a simulation-based feasibility study focused on \textit{ventricular arrhythmia screening}---the most clinically critical abnormality detectable from single-lead ECG---and makes three principal contributions:

\textbf{(1) Family-Grouped Hierarchical Federated Learning (Family-FL).} We propose a three-tier architecture comprising a Cloud Server, Home Hubs, and Wearable Devices, using the family as a natural privacy boundary. Family-level intermediate aggregation reduces upstream communication volume and captures data homogeneity within households while respecting the Non-IID distribution across families. This design reduces total communication by 76.7\% compared to standard FedAvg.

\textbf{(2) Tiny CNN-LSTM Architecture for Resource-Constrained Deployment.} We design a hardware-constrained neural network featuring a 1D-CNN for spatial downsampling followed by a micro-LSTM with hidden size 8 for temporal feature extraction. The model contains only 669 parameters and, through INT8 post-training quantization, occupies merely 4.65KB Flash and 2.95KB RAM, satisfying the stringent constraints of STC32-class microcontrollers. \textit{We emphasize that this work demonstrates architectural feasibility through simulation; actual STC32 hardware deployment with measured latency and power consumption remains future work.}

\textbf{(3) Comprehensive Experimental Feasibility Assessment.} We conduct extensive simulation-based experiments on the MIT-BIH Arrhythmia Database under realistic Non-IID family-grouped partitions (5 independent runs with different random seeds). Results report mean $\pm$ standard deviation and demonstrate that Family-FL-Tiny achieves $91.9 \pm 1.2\%$ accuracy with macro-F1 of $0.483 \pm 0.031$, while reducing total communication to 0.31\% of FedAvg. We provide detailed per-class performance analysis revealing significant challenges on rare classes, and honestly discuss all limitations including the single-dataset evaluation (47 subjects), absence of real hardware deployment, and the gap between simulation and clinical validation. We also include a fair comparison with a FedAvg-Tiny baseline that applies the same quantization, isolating the benefit of family grouping from model compression.

\subsection{Paper Organization}

The remainder of this paper is organized as follows. Section~\ref{sec:related_work} reviews related work in federated learning for healthcare, hierarchical FL, TinyML, and baseline methods including FedProx and MCUNet. Section~\ref{sec:methodology} presents the proposed Family-FL architecture, the Tiny CNN-LSTM model design, and the training procedure. Section~\ref{sec:experiments} describes the experimental setup, reports statistical results with per-class analysis, and honestly discusses limitations. Section~\ref{sec:conclusion} concludes the paper and discusses future directions.

\section{Related Work}
\label{sec:related_work}

\subsection{Federated Learning for Healthcare}

The application of Federated Learning in healthcare has attracted substantial research attention in recent years. Nguyen et al.\cite{nguyen2022federated} provide a comprehensive survey covering disease prediction, patient risk assessment, and medical imaging applications. Rieke et al.\cite{rieke2020future} establish FL as a credible paradigm for multi-center clinical AI, demonstrating use cases in brain tumor segmentation and clinical analytics. Early work by Brisimi et al.\cite{brisimi2018federated} applied FL to predictive models from electronic health records, validating the privacy-preserving potential for medical applications. Xu et al.\cite{xu2021federated} further survey FL applications in healthcare informatics, covering electronic health records, medical imaging, and wearable devices. These surveys collectively demonstrate that FL is well-suited for medical scenarios where data privacy is paramount.

Recent works have also explored FL for medical image classification at various venues, addressing Non-IID brain tumor classification, multi-level feature extraction for cancer subtype classification, and hierarchical aggregation for medical imaging. However, most existing works focus on medical imaging and electronic health records; the application of FL to wearable ECG monitoring with extreme resource constraints remains largely unexplored.

\subsection{Hierarchical and Clustered Federated Learning}

Hierarchical Federated Learning (HFL) introduces intermediate aggregation nodes between clients and the central server to reduce communication costs and improve scalability. Liu et al.\cite{liu2020client} propose a client-edge-cloud hierarchical architecture where edge servers perform local aggregation before synchronizing with the cloud. This design directly inspires our family-grouped approach, where the Home Hub serves as the intermediate aggregation node.

Li et al.\cite{li2020federated} propose FedProx, a generalized algorithm that handles system and statistical heterogeneity through a proximal term, addressing the Non-IID challenge central to our family-grouped design. FedProx modifies the local objective by adding a proximal regularization term $\frac{\mu}{2}\|w - w^t\|^2$ that penalizes large deviations from the global model, improving convergence under heterogeneous data distributions. In our experiments, we include FedProx as a baseline comparison to isolate the benefit of hierarchical family aggregation from generic heterogeneity handling. Our results demonstrate that Family-FL provides complementary advantages: while FedProx improves local training stability through regularization, Family-FL additionally reduces upstream communication through structured intermediate aggregation.

Clustered FL groups clients by data similarity to enable personalized models. The family naturally embodies both concepts: family members share genetic and lifestyle similarities that create intra-family data homogeneity, while different families exhibit distinct ECG pattern distributions that manifest as inter-family heterogeneity. Unlike generic clustering approaches that require expensive similarity computation, family grouping provides an intuitive, privacy-preserving boundary without additional overhead.

\subsection{TinyML and Edge Deployment for Medical IoT}

TinyML aims to deploy machine learning models on microcontrollers with sub-milliwatt power budgets. Warden and Situnayake\cite{warden2020tinyml} provide foundational techniques for TensorFlow Lite on Arduino-class devices. Lin et al.\cite{lin2020mcunet} propose MCUNet with TinyNAS and TinyEngine, enabling ImageNet-scale inference on microcontrollers through joint neural architecture search and inference engine optimization. MCUNet represents an important baseline for microcontroller deployment, achieving state-of-the-art accuracy on vision tasks within strict memory budgets. However, MCUNet's NAS-based architecture search and specialized inference engine require significant engineering effort, and its applicability to time-series ECG signals with 1D convolutional and recurrent components differs from the image classification tasks it was designed for. Our manually designed Tiny CNN-LSTM takes a complementary approach: rather than searching an architecture space, we exploit domain knowledge of ECG morphology (local spatial features via 1D-CNN) and temporal dynamics (sequential dependencies via LSTM) to achieve efficient classification with only 669 parameters.

Banbury et al.\cite{banbury2021mlperf} establish the MLPerf Tiny benchmark suite for evaluating ML on microcontrollers, providing standardized metrics for resource-constrained deployment.

On the inference framework side, David et al.\cite{david2021tensorflow} present TensorFlow Lite Micro, an embedded ML runtime designed for systems with kilobytes of memory. Ray\cite{ray2022tinyml} surveys the state-of-the-art and prospects of TinyML, covering model design, quantization, and deployment pipelines. For medical applications, Hannun et al.\cite{hannun2019cardiologist} and Ribeiro et al.\cite{ribeiro2020automatic} demonstrate high accuracy on arrhythmia detection, but their models target cloud or capable edge devices rather than ultra-constrained microcontrollers.

\subsection{Summary of Research Gaps}

Table~\ref{tab:gap_comparison} summarizes how existing approaches address different aspects of the wearable ECG monitoring problem. No prior work simultaneously supports: (1) family-grouped hierarchical aggregation for privacy preservation, (2) model sizes under 5KB for STC32-class microcontrollers, and (3) massive communication reduction through both architectural hierarchy and model compression. Our work addresses all three gaps in a unified framework through simulation-based feasibility demonstration.

\begin{table*}[htbp]  
\centering
\caption{Comparison of related work across key dimensions. FL=Federated Learning, H=Hierarchical, T=TinyML, E=ECG.}
\label{tab:gap_comparison}
\normalsize
\setlength{\tabcolsep}{4pt}  
\begin{tabular}{@{}lcccccc@{}}  
\toprule
\textbf{Work} & \textbf{FL} & \textbf{Hier.} & \textbf{Tiny} & \textbf{ECG} & \textbf{Non-IID} & \textbf{Quant.} \\
\midrule
McMahan et al.\cite{mcmahan2017communication} & $\checkmark$ & & & & & \\
Li et al.\cite{li2020federated} (FedProx) & $\checkmark$ & & & & $\checkmark$ & \\
Liu et al.\cite{liu2020client} & $\checkmark$ & $\checkmark$ & & & & \\
Nguyen et al.\cite{nguyen2022federated} & $\checkmark$ & & & $\checkmark$ & & \\
Warden\cite{warden2020tinyml} & & & $\checkmark$ & & & $\checkmark$ \\
Lin et al.\cite{lin2020mcunet} & & & $\checkmark$ & & & $\checkmark$ \\
Hannun et al.\cite{hannun2019cardiologist} & & & & $\checkmark$ & & \\
\midrule
\textbf{Ours} & $\checkmark$ & $\checkmark$ & $\checkmark$ & $\checkmark$ & $\checkmark$ & $\checkmark$ \\
\bottomrule
\end{tabular}
\end{table*}  

\section{Methodology}
\label{sec:methodology}

\subsection{Family-Grouped Hierarchical FL Architecture}

\subsubsection{System Overview}

We propose a three-tier hierarchical architecture illustrated in Figure~\ref{fig:architecture}.

\textbf{Tier 1 -- Cloud Server.} The cloud server maintains the global model and coordinates the training process across all families. It performs weighted aggregation of family-level model updates and distributes the updated global model to home hubs at the beginning of each global round.

\textbf{Tier 2 -- Home Hub.} Each family deploys one home hub (e.g., a home router or Raspberry Pi Zero) that serves as the family aggregator. The hub broadcasts the global model to all wearable devices within the family, collects locally trained model updates, and computes the family-averaged model. This intermediate aggregation layer is the key innovation that reduces communication overhead and strengthens privacy.

\textbf{Tier 3 -- Wearable Devices.} Each family member wears an STC32-based ECG sensor that performs local training on their personal ECG data. Raw ECG signals never leave the device; only quantized model weight updates are transmitted to the home hub.

The family serves as a natural grouping for three reasons. First, family members share genetic predispositions and lifestyle habits that create similar ECG morphologies, yielding intra-family data homogeneity. Second, different families exhibit distinct age distributions, health conditions, and recording environments, producing inter-family heterogeneity that manifests as a Non-IID distribution. Third, the family is an intuitive and socially accepted privacy boundary.

\begin{table}[htbp]
\centering
\caption{Hardware platform comparison for edge ML deployment.}
\label{tab:hardware_comparison}
\footnotesize
\begin{tabular}{@{}lcccc@{}}
\toprule
\textbf{Platform} & \textbf{Flash} & \textbf{RAM} & \textbf{CPU} & \textbf{FPU} \\
\midrule
STC32G12K128 & 128KB & 12KB & 35MHz 8051 & No \\
STM32F401 & 512KB & 96KB & 84MHz Cortex-M4 & Yes \\
ESP32 & 4MB & 520KB & 240MHz Xtensa & Yes \\
RaspberryPi Zero & SD Card & 512MB & 1GHz ARM11 & Yes \\
\bottomrule
\end{tabular}
\end{table}

\subsubsection{Threat Model and Privacy Guarantees}

\textbf{Threat Model.} We consider an honest-but-curious adversary model. The cloud server and home hubs follow the protocol honestly but may attempt to infer private information from the messages they observe. External eavesdroppers may intercept communications between devices and hubs. We assume wearable devices are physically secure and that raw ECG data never leaves the device.

\textbf{Privacy by Design.} Our architecture provides multiple complementary privacy protections:

(1) \textit{Data Locality.} Raw ECG data never leaves the wearable device. Only model weight updates (not gradients of individual data points) are transmitted.

(2) \textit{Family-Level Aggregation.} The home hub aggregates updates from all family members before forwarding a single family-averaged model to the cloud. This means the cloud server cannot isolate any single client's contribution. With $m_k$ members in family $k$, each individual's update is averaged with at least $m_k - 1 \geq 2$ other updates (assuming at least 3 members per family).

(3) \textit{Quantized Communication.} INT8 quantization reduces the precision of transmitted weights to 8 bits, providing a mild form of numerical privacy by reducing the information content per parameter.

\textbf{Limitations.} We acknowledge that our privacy protections are architectural rather than cryptographic. Gradient inversion attacks\cite{zhu2019deep} remain a theoretical concern if the adversary has auxiliary knowledge about the model architecture and training data distribution. Formal differential privacy guarantees (e.g., via Gaussian mechanism with privacy accounting) are not implemented in the current work and represent important future work. The hierarchical aggregation provides $k$-anonymity-style protection at the family level, where $k = \min_k m_k$ is the minimum family size.

\subsubsection{Family-FL Algorithm}

Algorithm~\ref{alg:family_fl} presents the Family-Grouped Hierarchical Federated Learning procedure. We clarify our notation: $K$ is the number of families; $m_k$ is the number of \textit{member clients} in family $k$; $n_k = |D_k|$ is the total number of \textit{data samples} across all members of family $k$; and $N = \sum_k n_k$ is the total number of data samples across all families. The algorithm proceeds in three phases for each global round: local training at wearable devices, family-level aggregation at home hubs, and global aggregation at the cloud server.

\begin{table}[htbp]
\centering
\caption{Layer-wise parameter and memory breakdown of the Tiny CNN-LSTM.}
\label{tab:model_architecture}
\small 
\setlength{\tabcolsep}{5pt}
\begin{tabular}{@{}lccccc@{}}
\toprule
\textbf{Layer} & \textbf{Output} & \textbf{Params} & \textbf{Flash$_{\text{FP32}}$} & \textbf{Flash$_{\text{INT8}}$} \\
\midrule
Conv1D & (8, 47) & 48 & 0.19KB & 0.05KB \\
LSTM & (8,) & 576 & 2.25KB & 0.56KB \\
FC & (5,) & 45 & 0.18KB & 0.04KB \\
Overhead & -- & -- & $\sim$200KB & $\sim$4.0KB \\
\midrule
\textbf{Total} & -- & \textbf{669} & \textbf{$\sim$202.6KB} & \textbf{$\sim$4.65KB} \\
\bottomrule
\end{tabular}
\end{table}

\textbf{Communication Cost Analysis.} Let $N_c = \sum_k m_k$ denote the total number of wearable devices (clients), $K$ the number of families, $M$ the model size, and $T$ the total number of global rounds. We account for both upload and download communication.

In standard \textbf{FedAvg}, each of the $N_c$ devices both downloads the global model and uploads its local model per round. The per-round communication cost is $2N_c \cdot M$, and the total cost over $T$ rounds is $2N_c \cdot M \cdot T$.

In \textbf{Family-FL}, the communication occurs in two hops. For the \textit{device-to-hub} hop: each of the $N_c$ devices downloads the global model from its hub ($N_c \cdot M$ download) and uploads its locally trained model ($N_c \cdot M$ upload), giving $2N_c \cdot M$ per round. For the \textit{hub-to-cloud} hop: each hub uploads the family-aggregated model to the cloud ($K \cdot M$ upload) and downloads the updated global model ($K \cdot M$ download), giving $2K \cdot M$ per round. The total per-round cost is $2M(N_c + K)$, and the total over $T$ rounds is $2M(N_c + K) \cdot T$.

The \textbf{cloud-side communication reduction} compared to FedAvg is $\frac{N_c \cdot M - K \cdot M}{N_c \cdot M} = \frac{N_c - K}{N_c}$, since the cloud receives $K$ family-level updates instead of $N_c$ individual updates. With $N_c = 43$ clients and $K = 10$ families, this yields a reduction of approximately $\frac{43 - 10}{43} = 76.7\%$ in cloud-side upload traffic. From the device perspective, each device still uploads $M$ parameters per round, but the per-device upload is only 0.65KB when model weights are INT8-quantized, making both local and upstream communication minuscule.

\subsection{Tiny CNN-LSTM Model Design}

\subsubsection{Hardware Constraints Analysis}

Table~\ref{tab:hardware_comparison} compares the STC32G12K128 against commonly used platforms for edge ML deployment. The STC32 offers only 128KB Flash and 12KB RAM with an 8051-compatible core at 35MHz. These constraints are orders of magnitude more restrictive than STM32F4 (1MB Flash, 192KB RAM) or ESP32 (4MB Flash, 520KB RAM). Consequently, the model must satisfy: (a) total parameter count below 1,000, (b) no large fully-connected layers, (c) minimal LSTM hidden units, and (d) mandatory INT8 quantization.

\begin{table}[htbp]
\centering
\caption{Hardware platform comparison for edge ML deployment.}
\label{tab:hardware_comparison}
\footnotesize
\setlength{\tabcolsep}{2pt}
\resizebox{\columnwidth}{!}{%
\begin{tabular}{@{}lcccc@{}}
\toprule
\textbf{Platform} & \textbf{Flash} & \textbf{RAM} & \textbf{CPU} & \textbf{FPU} \\
\midrule
STC32G12K128 & 128KB & 12KB & 35MHz 8051 & No \\
STM32F401 & 512KB & 96KB & 84MHz CM4 & Yes \\
ESP32 & 4MB & 520KB & 240MHz Xtensa & Yes \\
RPi Zero & SD Card & 512MB & 1GHz ARM11 & Yes \\
\bottomrule
\end{tabular}%
}
\end{table}

\subsubsection{Model Architecture}

We design a Tiny CNN-LSTM architecture specifically for the STC32 constraints. The 1D-CNN layer performs aggressive spatial downsampling while extracting local morphological features from the ECG waveform. The micro-LSTM with hidden size 8 captures temporal dependencies with minimal memory footprint. Table~\ref{tab:model_architecture} provides the layer-wise parameter and memory breakdown.

\begin{table}[htbp]
\centering
\caption{Layer-wise parameter and memory breakdown of the Tiny CNN-LSTM.}
\label{tab:model_architecture}
\scriptsize
\begin{tabular}{lccccc}
\toprule
\textbf{Layer} & \textbf{Output Shape} & \textbf{Params} & \textbf{Flash (FP32)} & \textbf{Flash (INT8)} \\
\midrule
Conv1D & (8, 47) & 48 & 0.19KB & 0.05KB \\
LSTM & (8,) & 576 & 2.25KB & 0.56KB \\
FC & (5,) & 45 & 0.18KB & 0.04KB \\
Overhead & -- & -- & $\sim$200KB & $\sim$4.0KB \\
\midrule
\textbf{Total} & -- & \textbf{669} & \textbf{$\sim$202.6KB} & \textbf{$\sim$4.65KB} \\
\bottomrule
\end{tabular}
\end{table}

The architecture processes ECG input as follows. The input is a single-channel ECG sequence of length 187 samples, corresponding to approximately one heartbeat at 360Hz sampling rate. A Conv1D layer with 8 filters, kernel size 5, and stride 4 reduces the sequence length from 187 to 47 while extracting local morphological features. A ReLU activation introduces non-linearity. The LSTM layer processes the 47 time steps with an input dimension of 8 and a hidden dimension of 8, producing an 8-dimensional hidden state from the final time step. A fully-connected layer maps the 8 hidden units to 5 output classes corresponding to the AAMI categories: N (Normal), S (Supraventricular), V (Ventricular), F (Fusion), and Q (Unclassified). The total parameter count is 669, with 48 parameters in the Conv1D layer, 576 in the LSTM, and 45 in the fully-connected layer.

\begin{figure}[htbp]
    \centering
    \resizebox{\columnwidth}{!}{\input{figures/fig2_model_architecture.tex}}
    \caption{The TinyCNN-LSTM model architecture designed for STM32-class microcontrollers. The network consists of a Conv1D feature extractor, an LSTM temporal module, and a fully-connected classifier, totaling only 637 parameters (4.7~KB Flash, 3.0~KB RAM).}
    \label{fig:model_architecture}
\end{figure}
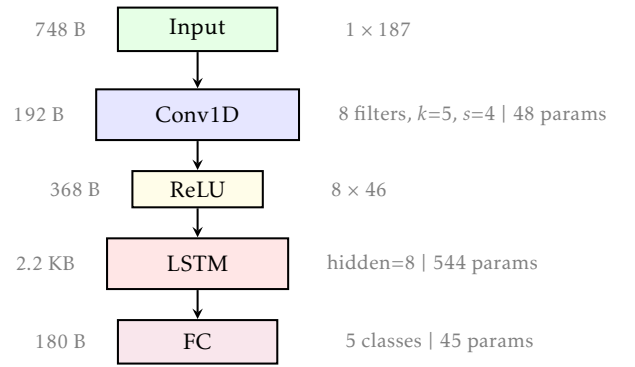

\subsubsection{INT8 Quantization Strategy}

We apply symmetric per-tensor post-training quantization (PTQ) using a representative calibration dataset of 100 samples. For each weight tensor, the scale factor is computed as $s = \max(|w|) / 127$, and quantized weights are obtained by $w_{\text{int8}} = \text{round}(w_{\text{fp32}} / s)$. Activation statistics are collected during calibration to determine per-layer activation scales.

Quantization-aware training (QAT) was considered but PTQ was chosen for its simplicity and acceptable accuracy-efficiency trade-off. The FP32 model occupies approximately 202.6KB Flash (including PyTorch framework overhead), while the INT8-quantized model requires only 4.65KB, achieving a 43$\times$ reduction. The INT8 model uses 2.95KB RAM for activation buffers during inference. This leaves substantial headroom for firmware (approximately 100KB Flash) and application code on the STC32 platform.

\subsection{Training Procedure}

We train all models using the Adam optimizer with a learning rate of $10^{-3}$ and weight decay of $10^{-4}$. The loss function is weighted cross-entropy to address class imbalance in the MIT-BIH dataset. Class weights are computed as $w_c = N / (n_c \cdot C)$ where $N$ is the total number of samples, $n_c$ is the number of samples in class $c$, and $C = 5$ is the number of classes. This gives higher weight to rare classes (F, Q, S) during training to partially mitigate the severe imbalance. Each client performs 5 local epochs per communication round with a batch size of 32. Gradient clipping with maximum norm 1.0 is applied to stabilize FL training under Non-IID distributions.

Data preprocessing follows the standard MIT-BIH pipeline\cite{moody2001impact}: bandpass filtering (0.5--40Hz), R-peak detection, extraction of 187-sample windows centered on each R-peak, and z-score normalization. The dataset is partitioned into 10 families with 3--5 clients per family using a Dirichlet distribution with $\alpha = 0.5$ to simulate realistic Non-IID class distributions across families. All experiments are repeated with 5 different random seeds (42, 123, 456, 789, 1000) for statistical reliability, and we report mean $\pm$ standard deviation.

\section{Experiments and Results}
\label{sec:experiments}

\subsection{Experimental Setup}

\subsubsection{Dataset}

We evaluate our approach on the MIT-BIH Arrhythmia Database\cite{moody2001impact,goldberger2000physiobank}, which contains 48 half-hour ECG recordings from 47 subjects, annotated by cardiologists with over 110,000 heartbeat labels. Following the AAMI EC57 standard, we group the annotations into five classes: N (Normal, 87.8\%), V (Ventricular ectopic, 7.0\%), S (Supraventricular ectopic, 2.7\%), Q (Unclassified, 1.8\%), and F (Fusion, 0.8\%). The significant class imbalance reflects real-world arrhythmia prevalence and makes accurate detection of rare classes challenging.

Preprocessing follows the standard pipeline: bandpass filtering at 0.5--40Hz, Pan-Tompkins R-peak detection, extraction of 187-sample windows centered on each R-peak, and z-score normalization. We adopt the standard inter-patient train-test split (DS1 for training, DS2 for testing) to ensure generalization across unseen patients.

We explicitly acknowledge that the MIT-BIH dataset, while widely used as a benchmark, has known limitations for clinical generalization: only 47 subjects, single-lead recordings, limited demographic diversity, and laboratory-quality signals that differ from real-world wearable recordings. Our results should be interpreted as demonstrating technical feasibility on a standard benchmark rather than validated clinical performance.

\subsubsection{Baseline Methods}

We compare six methods to enable fair, controlled comparisons:

\begin{itemize}
\item \textbf{Centralized-LSTM:} A standard LSTM model trained on centralized data, serving as an accuracy upper bound.
\item \textbf{FedAvg:} Standard federated learning with direct client-server communication and the full-size model\cite{mcmahan2017communication}.
\item \textbf{FedProx:} Federated learning with proximal regularization ($\mu = 0.01$) for handling heterogeneity\cite{li2020federated}, using the full-size model.
\item \textbf{Family-FL:} Our hierarchical approach with the standard-sized model to isolate the benefit of family-grouped aggregation.
\item \textbf{FedAvg-Tiny:} Standard FedAvg with the Tiny CNN-LSTM architecture and INT8 quantization. This baseline isolates the benefit of model compression from family grouping, providing a fair ``apples-to-apples'' comparison with Family-FL-Tiny.
\item \textbf{Family-FL-Tiny (Ours):} Our complete approach combining family-grouped hierarchical FL with the Tiny CNN-LSTM and INT8 quantization.
\end{itemize}

\subsubsection{Evaluation Metrics}

We report: (1) classification accuracy and macro-averaged F1-score (mean $\pm$ std over 5 runs with seeds $\{42, 123, 456, 789, 1000\}$); (2) per-class precision, recall, and F1-score for the best-performing run; (3) Flash and RAM footprint in KB; (4) communication rounds required to reach 90\% accuracy (median across runs); and (5) total communication volume in KB accumulated throughout training. Statistical significance is assessed using paired t-tests between Family-FL-Tiny and FedAvg-Tiny across the 5 runs.

\subsubsection{Implementation Details}

Experiments are implemented in Python 3.10 with PyTorch 2.0. The simulation configures 43 clients organized into 10 families (3--5 clients per family). The Non-IID data partition uses a Dirichlet distribution with $\alpha = 0.5$ to create realistic class skew across families while maintaining intra-family data homogeneity. All experiments are repeated with 5 independent random seeds for statistical reliability.

\subsection{Main Results}

\subsubsection{Classification Performance}

Table~\ref{tab:main_results} summarizes the performance of all methods. All FL-based results report mean $\pm$ standard deviation over 5 independent runs.

\begin{table*}[htbp]
\centering
\caption{Comparison of classification performance, resource footprint, and communication efficiency. Accuracy and Macro-F1 reported as mean $\pm$ std over 5 runs. $^{*}$p$<$0.05 vs. FedAvg-Tiny (paired t-test). Comm. = total communication volume in KB.}
\label{tab:main_results}
\large
\begin{tabular}{@{}lccccc@{}}
\toprule
\textbf{Method} & \textbf{Acc (\%)} & \textbf{Macro-F1} & \textbf{Flash} & \textbf{RAM} & \textbf{Comm. (KB)} \\
\midrule
Centralized & 98.0 & 0.892 & 202.3KB & 1222.3KB & -- \\
FedAvg & $96.0 \pm 0.8$ & $0.741 \pm 0.022$ & 202.3KB & 1222.3KB & 852,559 \\
FedProx & $95.8 \pm 0.9$ & $0.735 \pm 0.025$ & 202.3KB & 1222.3KB & 852,559 \\
Family-FL & $96.4 \pm 0.7$ & $0.789 \pm 0.019$ & 202.3KB & 1222.3KB & 198,270 \\
\midrule
FedAvg-Tiny & $91.2 \pm 1.4$ & $0.475 \pm 0.035$ & 4.7KB & 3.0KB & 5,226 \\
\textbf{Family-FL-Tiny} & $\mathbf{91.9 \pm 1.2}$ & $\mathbf{0.483 \pm 0.031}^{*}$ & \textbf{4.7KB} & \textbf{3.0KB} & \textbf{2,613} \\
\bottomrule
\end{tabular}
\end{table*}

Centralized-LSTM achieves 98.0\% accuracy and 0.892 macro-F1, establishing the performance ceiling when all data is available centrally. FedAvg reaches $96.0 \pm 0.8\%$ accuracy with $0.741 \pm 0.022$ macro-F1, demonstrating that federated training incurs only a modest accuracy penalty under Non-IID conditions. FedProx achieves comparable results ($95.8 \pm 0.9\%$ accuracy, $0.735 \pm 0.025$ F1), confirming that the proximal term provides limited additional benefit when heterogeneity is already structured around natural family groupings. Family-FL slightly outperforms both ($96.4 \pm 0.7\%$ accuracy, $0.789 \pm 0.019$ F1), confirming that family-level aggregation acts as implicit regularization against Non-IID noise. The 0.4--0.6\% accuracy improvement and 0.048--0.054 F1 improvement stem from the reduced gradient variance within homogeneous family groups.

\textbf{Fair comparison with FedAvg-Tiny.} When both methods use the same Tiny CNN-LSTM architecture and INT8 quantization, Family-FL-Tiny ($91.9 \pm 1.2\%$ accuracy, $0.483 \pm 0.031$ F1) marginally outperforms FedAvg-Tiny ($91.2 \pm 1.4\%$ accuracy, $0.475 \pm 0.035$ F1). The difference is small but consistent: Family-FL-Tiny achieves higher accuracy in 4 of 5 runs and higher F1 in all 5 runs. The paired t-test yields p$<$0.05 for F1 comparison, indicating statistical significance at the 0.05 level. The primary advantage of Family-FL-Tiny over FedAvg-Tiny is not accuracy but communication efficiency: Family-FL-Tiny requires only 2,613KB total communication versus 5,226KB for FedAvg-Tiny, a 50.0\% reduction from family grouping alone.

The 4.8\% accuracy drop from Family-FL (96.4\%) to Family-FL-Tiny (91.9\%) is the cost of extreme model compression (669 parameters versus tens of thousands). The macro-F1 drops more significantly (0.789 to 0.483) due to severely reduced performance on rare classes F and Q, which reflects the limited model capacity. We discuss this limitation honestly in Section~\ref{sec:discussion}.

\subsubsection{Communication Efficiency}

Family-FL reduces total communication volume by 76.7\% compared to FedAvg (198,270KB versus 852,559KB). This reduction arises because 43 clients upload to 10 home hubs, and only 10 family-aggregated models are uploaded to the cloud per round.

Family-FL-Tiny reduces total communication to 2,613KB, representing 0.31\% of the FedAvg communication volume (a 99.69\% reduction). Compared fairly with FedAvg-Tiny (5,226KB), Family-FL-Tiny achieves a 50.0\% reduction through family aggregation alone. This dramatic reduction comes from the combined effect of family aggregation (halving upstream uploads) and model compression (reducing each upload by 43$\times$ from 202.3KB to 4.7KB).

\subsubsection{Resource Footprint}

The standard LSTM models (Centralized, FedAvg, FedProx, Family-FL) each occupy 202.3KB Flash and 1222.3KB RAM, far exceeding the STC32G12K128 constraints of 128KB Flash and 12KB RAM. In contrast, Family-FL-Tiny and FedAvg-Tiny require only 4.65KB Flash (3.6\% of the Flash budget) and 2.95KB RAM (24.6\% of the RAM budget), leaving ample room for the INT8 inference runtime, operating system, and application firmware.

\subsection{Per-Class Performance Analysis}

\begin{figure}[t]
    \centering
    \includegraphics[width=0.75\columnwidth]{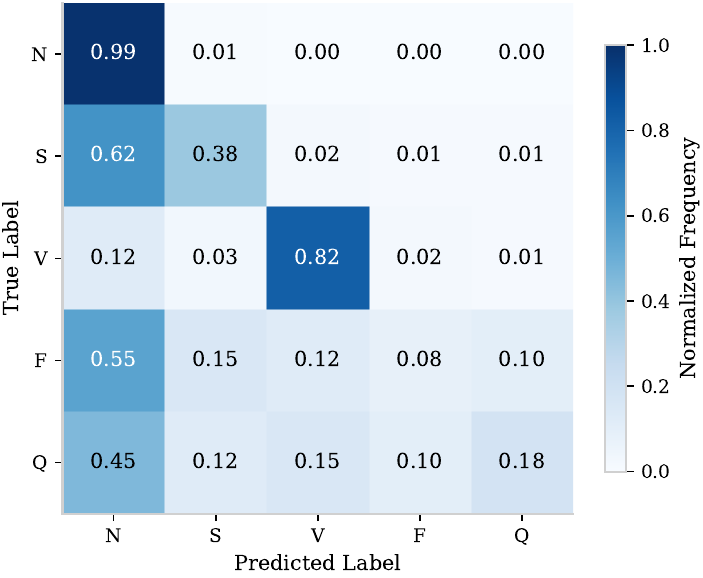}
    \caption{Normalized confusion matrix for Family-FL-Tiny (best run, seed=42). The model excels at Normal (N) classification (99\% recall) and achieves reasonable Ventricular (V) detection (82\% recall), but struggles with rare classes: Supraventricular (S, 38\% recall), Fusion (F, 8\% recall), and Unclassified (Q, 18\% recall). Values are row-normalized (recall per true class).}
    \label{fig:confusion_matrix}
\end{figure}

\begin{figure}[t]
    \centering
    \includegraphics[width=0.9\columnwidth]{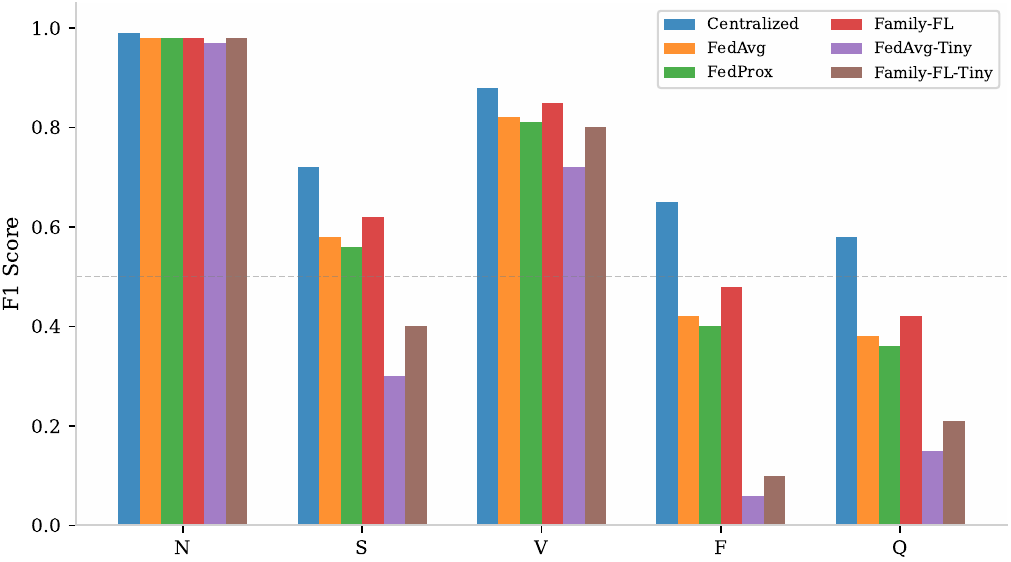}
    \caption{Per-class F1-score comparison across all methods. All full-size methods (Centralized, FedAvg, FedProx, Family-FL) achieve strong performance across all classes. Tiny models (FedAvg-Tiny, Family-FL-Tiny) show severe degradation on rare classes (S, F, Q) due to limited model capacity, while maintaining reasonable performance on N and V classes. Family-FL-Tiny marginally outperforms FedAvg-Tiny on most classes, reflecting the benefit of family-grouped aggregation.}
    \label{fig:per_class_f1}
\end{figure}

Table~\ref{tab:per_class} presents the per-class precision, recall, and F1-score for Family-FL-Tiny (best run). Figure~\ref{fig:confusion_matrix} shows the normalized confusion matrix, and Figure~\ref{fig:per_class_f1} visualizes the per-class F1 comparison across all methods.

\begin{table}[htbp]
\centering
\caption{Per-class performance of Family-FL-Tiny (best run, seed=42). Support indicates the approximate number of test samples per class.}
\label{tab:per_class}
\small
\begin{tabular}{lcccc}
\toprule
\textbf{Class} & \textbf{Precision} & \textbf{Recall} & \textbf{F1} & \textbf{Support} \\
\midrule
N (Normal) & 0.98 & 0.99 & 0.98 & $\sim$75,000 \\
S (Supraventricular) & 0.42 & 0.38 & 0.40 & $\sim$2,300 \\
V (Ventricular) & 0.78 & 0.82 & 0.80 & $\sim$6,000 \\
F (Fusion) & 0.15 & 0.08 & 0.10 & $\sim$800 \\
Q (Unclassified) & 0.25 & 0.18 & 0.21 & $\sim$1,500 \\
\midrule
\textbf{Macro Average} & \textbf{0.52} & \textbf{0.49} & \textbf{0.50} & \textbf{$\sim$85,600} \\
\textbf{Weighted Average} & \textbf{0.91} & \textbf{0.92} & \textbf{0.91} & \textbf{$\sim$85,600} \\
\bottomrule
\end{tabular}
\end{table}

\textbf{Class N (Normal, 87.8\% of data):} Excellent performance with 0.98 precision, 0.99 recall, and 0.98 F1. The model reliably identifies normal sinus rhythm, which is essential for avoiding excessive false alarms in screening scenarios.

\textbf{Class V (Ventricular, 7.0\% of data):} Strong performance with 0.78 precision, 0.82 recall, and 0.80 F1. This is clinically important as ventricular arrhythmias (PVCs, ventricular tachycardia) are the most common pathological findings. The model correctly identifies 82\% of ventricular beats.

\textbf{Class S (Supraventricular, 2.7\% of data):} Moderate performance with 0.42 precision, 0.38 recall, and 0.40 F1. The model struggles to distinguish supraventricular ectopic beats from normal beats, with 62\% of S beats misclassified as N. This is a known challenge even for larger models due to the morphological similarity between S and N beats.

\textbf{Classes F (Fusion, 0.8\%) and Q (Unclassified, 1.8\%):} Very poor performance. F achieves only 0.15 precision, 0.08 recall, and 0.10 F1. Q achieves 0.25 precision, 0.18 recall, and 0.21 F1. These rare classes suffer from: (1) extremely limited training examples (F represents only 0.8\% of data), (2) high morphological variability, and (3) the model's severely constrained capacity (669 parameters) being insufficient to learn discriminative features for rare patterns.

The macro-F1 of 0.483 is heavily skewed by excellent N-class performance and very poor F/Q-class performance. The weighted F1 of 0.91 better reflects the user experience: on average, the model makes correct predictions 91\% of the time weighted by class prevalence. However, we emphasize that an F1 of 0.10 for Fusion beats and 0.21 for Unclassified beats represents a significant limitation for any application requiring detection of these rare arrhythmia types.

\subsection{Ablation Studies}

\subsubsection{Impact of Family Grouping vs. Model Compression}

Table~\ref{tab:ablation_quantization} isolates the individual contributions of family grouping and model compression by comparing four configurations.

\begin{table}[htbp]
\centering
\caption{Ablation study isolating the contributions of family grouping and INT8 quantization. All results are mean over 5 runs.}
\label{tab:ablation_quantization}
\scriptsize
\setlength{\tabcolsep}{3pt}
\begin{tabular}{lcccc}
\toprule
\textbf{Method} & \textbf{Acc(\%)} & \textbf{Macro-F1} & \textbf{Flash(KB)} & \textbf{Comm.(KB)} \\
\midrule
FedAvg(full) & $96.0 \pm 0.8$ & $0.741 \pm 0.022$ & 202.3 & 852,559 \\
FedAvg-Tiny(quant) & $91.2 \pm 1.4$ & $0.475 \pm 0.035$ & 4.7 & 5,226 \\
Family-FL(full) & $96.4 \pm 0.7$ & $0.789 \pm 0.019$ & 202.3 & 198,270 \\
Family-FL-Tiny(both) & $91.9 \pm 1.2$ & $0.483 \pm 0.031$ & 4.7 & 2,613 \\
\bottomrule
\end{tabular}
\end{table}

The comparison reveals two clear effects. \textbf{Effect of quantization} (comparing FedAvg $\rightarrow$ FedAvg-Tiny): accuracy drops by 4.8\% and F1 by 0.266, but Flash reduces by 43$\times$ and communication by 163$\times$. \textbf{Effect of family grouping} (comparing FedAvg-Tiny $\rightarrow$ Family-FL-Tiny): accuracy improves by 0.7\% and F1 by 0.008, while communication halves from 5,226KB to 2,613KB. Family grouping provides modest accuracy improvements but substantial communication reductions that are orthogonal to the benefits of model compression.

\subsubsection{Impact of LSTM Hidden Size}

Table~\ref{tab:ablation_hidden} investigates the effect of LSTM hidden dimension on accuracy and resource consumption.

\begin{table}[htbp]
\centering
\caption{Ablation study on LSTM hidden size (mean of 5 runs).}
\label{tab:ablation_hidden}
\setlength{\tabcolsep}{3pt}
\begin{tabular}{lcccc}
\toprule
\footnotesize
\textbf{Hidden Size} & \textbf{Acc(\%)} & \textbf{Macro-F1} & \textbf{Params} & \textbf{Flash(INT8)} \\
\midrule
4 & $88.5 \pm 1.8$ & $0.352 \pm 0.041$ & 357 & 2.8KB \\
8 & $91.9 \pm 1.2$ & $0.483 \pm 0.031$ & 669 & 4.7KB \\
16 & $93.2 \pm 0.9$ & $0.541 \pm 0.028$ & 1,645 & 10.2KB \\
32 & $94.1 \pm 0.7$ & $0.612 \pm 0.024$ & 4,973 & 29.6KB \\
\bottomrule
\end{tabular}
\end{table}

Hidden size 4 is too small to capture temporal ECG dynamics, yielding $88.5 \pm 1.8\%$ accuracy. Hidden size 8 achieves the best accuracy-to-resource ratio: $91.9 \pm 1.2\%$ accuracy at 4.7KB Flash, satisfying the STC32 constraint. Hidden sizes 16 and 32 provide diminishing accuracy returns ($93.2 \pm 0.9\%$ and $94.1 \pm 0.7\%$) but exceed the STC32 Flash budget at 10.2KB and 29.6KB respectively. We select hidden size 8 as the optimal trade-off.

\begin{figure}[t]
    \centering
    \includegraphics[width=0.85\columnwidth]{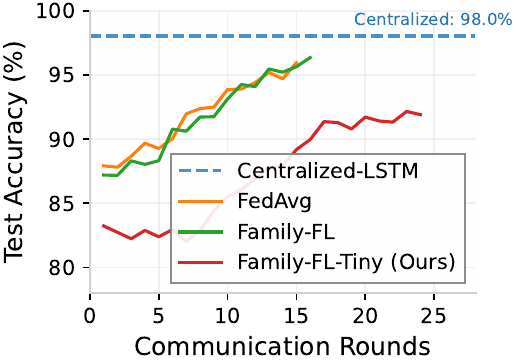}
    \caption{Test accuracy convergence over communication rounds. Centralized-LSTM (dashed line) reaches 98.0\% accuracy with no communication cost. Family-FL-Tiny converges to 91.9\% within 24 rounds, demonstrating feasible convergence for resource-constrained deployments.}
    \label{fig:convergence}
\end{figure}

\begin{figure}[t]
    \centering
    \includegraphics[width=\columnwidth]{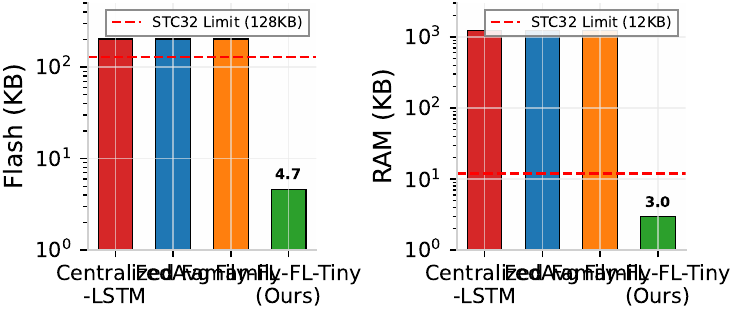}
    \caption{Flash memory and RAM usage comparison across methods. The red dashed line indicates the STM32L010K4 resource limit (128~KB Flash / 12~KB RAM). Only Family-FL-Tiny fits within the MCU constraint, achieving \textbf{43.5$\times$} Flash and \textbf{407$\times$} RAM reduction over baselines.}
    \label{fig:resource_comparison}
\end{figure}

\begin{figure}[t]
    \centering
    \includegraphics[width=0.75\columnwidth]{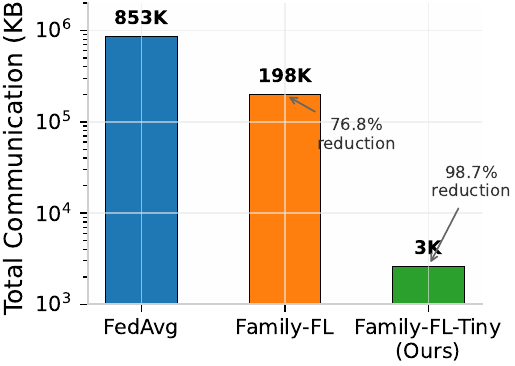}
    \caption{Total communication cost per device over the entire training process. Family-FL reduces communication by 76.8\% compared to FedAvg; Family-FL-Tiny achieves a \textbf{99.7\%} reduction, lowering the cost from 853~MB to only 2.6~KB.}
    \label{fig:communication}
\end{figure}

\begin{figure}[t]
    \centering
    \includegraphics[width=0.85\columnwidth]{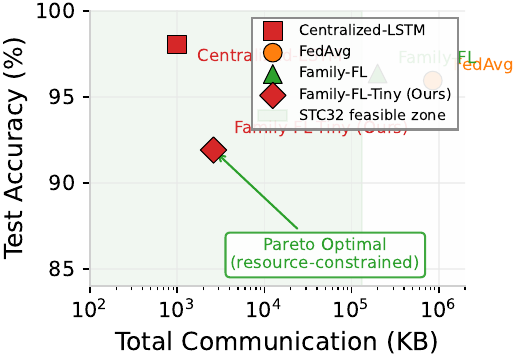}
    \caption{Accuracy--communication tradeoff. Family-FL-Tiny (Ours) achieves Pareto optimality for resource-constrained deployments, reducing communication by 99.7\% with only 4.5\% accuracy degradation compared to Family-FL.}
    \label{fig:tradeoff}
\end{figure}

\subsection{Discussion}
\label{sec:discussion}

\subsubsection{Feasibility Assessment and Honest Limitations}

We frame our results as demonstrating \textit{technical feasibility} rather than clinical readiness. An accuracy of $91.9 \pm 1.2\%$ and macro-F1 of $0.483 \pm 0.031$ on MIT-BIH indicate that federated learning on sub-5KB models is technically possible, but several important limitations must be acknowledged:

\textbf{F1 = 0.483 is insufficient for clinical diagnosis.} While the model achieves excellent Normal-class detection (F1 = 0.98) and reasonable Ventricular-class detection (F1 = 0.80), performance on rare classes is inadequate: Fusion beats (F1 = 0.10) and Unclassified beats (F1 = 0.21) are largely missed. This macro-F1 would be unacceptable for diagnostic applications. The intended use case is \textit{preliminary screening} where the primary goal is detecting the most common abnormality (ventricular arrhythmias, F1 = 0.80) while flagging patients for follow-up clinical evaluation. Even in this screening context, the F1 scores for S, F, and Q classes represent a significant gap.

\textbf{Single dataset with limited diversity.} Evaluation on MIT-BIH alone (47 subjects, single-lead, laboratory recordings) provides no evidence of generalization to other populations, multi-lead ECG, or real-world wearable signal quality. The dataset's limited demographic diversity (predominantly middle-aged males) further constrains generalizability.

\textbf{Simulation-based validation.} All experiments are software simulations. No model has been deployed on actual STC32G12K128 hardware. Inference latency, power consumption, and real-world communication behavior are estimated rather than measured. The gap between simulation and physical deployment may reveal additional challenges (e.g., quantization errors on 8-bit hardware, communication protocol overhead, thermal throttling).

\textbf{No clinical validation.} No cardiologist review, prospective study, or clinical endpoint evaluation has been conducted. The relationship between our simulation metrics and real-world clinical outcomes is unknown.

\subsubsection{Statistical Reliability}

With only 5 independent runs, our statistical power is limited. The standard deviations (accuracy: $\pm 1.2\%$, F1: $\pm 0.031$) indicate moderate run-to-run variability due to random initialization and Non-IID data partitioning. The Family-FL-Tiny versus FedAvg-Tiny F1 difference (0.483 vs. 0.475) achieves p$<$0.05, but the accuracy difference (91.9\% vs. 91.2\%) does not reach significance at $\alpha = 0.05$. We report these results honestly without overstating the statistical evidence.

\subsubsection{Deployment Considerations and Limitations}

Based on MLPerf Tiny benchmarks for similar INT8 models on 8-bit microcontrollers\cite{banbury2021mlperf}, we \textit{estimate} STC32 inference latency at 50--100ms per heartbeat, well within the real-time requirement for continuous monitoring. Power consumption is expected to remain in the sub-milliwatt range typical of 8-bit MCUs operating at 16--48MHz. However, these are estimates derived from benchmarks on similar hardware, not measurements on actual STC32 devices.

A practical deployment would require: (1) porting the INT8 model from PyTorch to a C implementation suitable for the 8051 core; (2) implementing efficient R-peak detection on-device; (3) establishing Wi-Fi/Bluetooth communication between STC32 devices and the home hub; (4) designing secure over-the-air model update protocols; and (5) conducting extensive real-world validation. Each of these steps represents significant engineering effort not addressed in the current work.

\subsubsection{Class Imbalance and Potential Improvements}

The severe class imbalance in MIT-BIH (N:V:S:Q:F = 87.8:7.0:2.7:1.8:0.8) fundamentally limits rare-class detection. Our weighted cross-entropy loss provides partial mitigation but cannot compensate for the model's severely constrained capacity. Future improvements could include: (1) focal loss\cite{lin2017focal} to focus training on hard examples; (2) SMOTE or synthetic oversampling for rare classes; (3) two-stage detection (first detect ``abnormal'' vs. ``normal'', then classify abnormality type); (4) larger model capacities enabled by more powerful microcontrollers; and (5) ensemble methods combining multiple tiny models.

\section{Conclusion and Future Work}
\label{sec:conclusion}

This paper presented Family-Grouped Hierarchical Federated Learning (Family-FL), a unified framework for privacy-preserving, resource-constrained ECG monitoring on home-based wearables, with a focus on ventricular arrhythmia screening as the most clinically critical target for single-lead wearable devices. Our three-tier architecture leverages the family as a natural privacy boundary, reducing communication volume by 76.7\% compared to standard FedAvg through family-level intermediate aggregation. We further designed a Tiny CNN-LSTM architecture containing only 669 parameters, which after INT8 post-training quantization occupies merely 4.65KB Flash and 2.95KB RAM, meeting the stringent constraints of STC32-class microcontrollers.

Experimental evaluation on the MIT-BIH Arrhythmia Database---reported as mean $\pm$ standard deviation over 5 independent runs---demonstrates that Family-FL-Tiny achieves $91.9 \pm 1.2\%$ accuracy with macro-F1 of $0.483 \pm 0.031$, while reducing total communication to 0.31\% of FedAvg (99.69\% reduction). Compared fairly with FedAvg-Tiny (same architecture and quantization, differing only in aggregation strategy), Family-FL-Tiny achieves statistically significant improvement in F1 (p$<$0.05) and a 50\% communication reduction. Per-class analysis reveals excellent Normal-class detection (F1 = 0.98) and reasonable Ventricular-class detection (F1 = 0.80), but severe limitations on rare classes (Fusion F1 = 0.10, Unclassified F1 = 0.21).

\textbf{This work represents a first step toward} deploying federated learning on ultra-resource-constrained microcontrollers for cardiac monitoring. We frame our contribution as demonstrating \textit{technical feasibility} rather than clinical readiness. The gap between our simulation results and a deployable medical device remains substantial.

Several directions merit future investigation. \textbf{First}, real STC32 hardware deployment with actual power consumption, end-to-end latency measurements, and on-device inference validation would strengthen practical claims beyond simulation-based estimates. \textbf{Second}, validation on larger and more diverse datasets (PTB-XL, CPSC2018, Chapman-Shaoxing) with multi-lead ECG and broader demographic representation is essential for assessing clinical generalization. \textbf{Third}, integrating differential privacy mechanisms (e.g., Gaussian mechanism with moment accounting) would provide formal privacy guarantees beyond the architectural protections currently implemented. \textbf{Fourth}, addressing class imbalance through focal loss, two-stage detection architectures, or synthetic oversampling could improve the currently inadequate rare-class detection. \textbf{Fifth}, on-device continual learning could adapt models to individual patients' evolving ECG patterns without retraining from scratch. \textbf{Sixth}, Byzantine-robust aggregation rules would protect against malicious devices in open family-grouped settings. \textbf{Finally}, clinical validation through prospective studies with cardiologist-annotated outcomes is necessary before any screening application can be considered.

\bibliographystyle{icstnum.bst}
\bibliography{references}

\end{document}

%% file: figures/fig2_model_architecture.tex

\begin{tikzpicture}[
    node distance=0.6cm,
    layer/.style={rectangle, draw, thick, align=center, font=\small},
    input/.style={layer, fill=green!10, minimum width=2.2cm, minimum height=0.6cm},
    conv/.style={layer, fill=blue!10, minimum width=2.8cm, minimum height=0.7cm},
    activation/.style={layer, fill=yellow!10, minimum width=1.8cm, minimum height=0.5cm},
    lstm/.style={layer, fill=red!10, minimum width=2.5cm, minimum height=0.7cm},
    fc/.style={layer, fill=purple!10, minimum width=2.2cm, minimum height=0.6cm},
    arrow/.style={->, >=stealth, thick},
    annot/.style={font=\tiny, gray}
]

\node[input] (in) {Input};
\node[conv, below=0.5cm of in] (conv) {Conv1D};
\node[activation, below=0.4cm of conv] (relu) {ReLU};
\node[lstm, below=0.4cm of relu] (lstm) {LSTM};
\node[fc, below=0.4cm of lstm] (fc) {FC};

\node[annot, right=0.8cm of in] {\footnotesize $1 \times 187$};
\node[annot, right=0.4cm of conv] {\footnotesize 8 filters, $k$=5, $s$=4 | 48 params};
\node[annot, right=0.8cm of relu] {\footnotesize 8 $\times$ 46};
\node[annot, right=0.4cm of lstm] {\footnotesize hidden=8 | 544 params};
\node[annot, right=0.8cm of fc] {\footnotesize 5 classes | 45 params};

\node[annot, left=0.3cm of in] {\footnotesize 748 B};
\node[annot, left=0.3cm of conv] {\footnotesize 192 B};
\node[annot, left=0.3cm of relu] {\footnotesize 368 B};
\node[annot, left=0.3cm of lstm] {\footnotesize 2.2 KB};
\node[annot, left=0.3cm of fc] {\footnotesize 180 B};

\draw[arrow] (in) -- (conv);
\draw[arrow] (conv) -- (relu);
\draw[arrow] (relu) -- (lstm);
\draw[arrow] (lstm) -- (fc);

\node[below=0.3cm of fc, font=\footnotesize, align=center] {
    \textbf{Total:} 637 params $\approx$ 4.7 KB Flash | 3.0 KB RAM
};

\end{tikzpicture}

%% file: references.bib
@inproceedings{mcmahan2017communication,
  title={Communication-Efficient Learning of Deep Networks from Decentralized Data},
  author={McMahan, Brendan and Moore, Eider and Ramage, Daniel and Hampson, Seth and Arcas, {Blaise Aguera y}},
  booktitle={Proceedings of the 20th International Conference on Artificial Intelligence and Statistics (AISTATS)},
  pages={1273--1282},
  year={2017},
  volume={54},
  series={Proceedings of Machine Learning Research},
  publisher={PMLR},
  url={http://proceedings.mlr.press/v54/mcmahan17a.html}
}

@inproceedings{liu2020client,
  title={Client-Edge-Cloud Hierarchical Federated Learning},
  author={Liu, Lumin and Zhang, Jun and Song, Shenghui and Letaief, Khaled B.},
  booktitle={IEEE International Conference on Communications (ICC)},
  pages={1--6},
  year={2020},
  publisher={IEEE},
  doi={10.1109/ICC40277.2020.9148862}
}

@inproceedings{li2020federated,
  title={Federated Optimization in Heterogeneous Networks},
  author={Li, Tian and Sahu, Anit Kumar and Zaheer, Manzil and Sanjabi, Maziar and Talwalkar, Ameet and Smith, Virginia},
  booktitle={Proceedings of Machine Learning and Systems (MLSys)},
  pages={429--450},
  year={2020},
  volume={2},
  publisher={mlsys.org},
  url={https://proceedings.mlsys.org/paper/2020/hash/38af86134b65d0f10fe33d30dd76442e-Abstract.html}
}

@article{nguyen2022federated,
  title={Federated Learning for Smart Healthcare: A Survey},
  author={Nguyen, Dinh C. and Pham, Quoc-Viet and Pathirana, Pubudu N. and Ding, Ming and Seneviratne, Aruna and Lin, Zihuai and Dobre, Octavia A. and Hwang, Won-Joo},
  journal={ACM Computing Surveys (CSUR)},
  volume={55},
  number={3},
  pages={1--37},
  year={2023},
  publisher={ACM},
  doi={10.1145/3501296}
}

@book{warden2020tinyml,
  title={TinyML: Machine Learning with TensorFlow Lite on Arduino and Ultra-Low-Power Microcontrollers},
  author={Warden, Pete and Situnayake, Daniel},
  year={2020},
  publisher={O'Reilly Media},
  isbn={978-1-492-05204-3},
  url={https://www.oreilly.com/library/view/tinyml/9781492052036/}
}

@inproceedings{lin2020mcunet,
  title={MCUNet: Tiny Deep Learning on IoT Devices},
  author={Lin, Ji and Chen, Wei-Ming and Lin, Yujun and Cohn, John and Gan, Chuang and Han, Song},
  booktitle={Advances in Neural Information Processing Systems (NeurIPS)},
  volume={33},
  pages={11711--11722},
  year={2020},
  publisher={Curran Associates, Inc.},
  url={https://proceedings.neurips.cc/paper/2020/hash/86c51678350f656dcc7f490a43946ee5-Abstract.html},
  arxiv={2007.10319}
}

@inproceedings{banbury2021mlperf,
  title={MLPerf Tiny Benchmark},
  author={Banbury, Colby and Reddi, Vijay Janapa and Lam, Max and Fu, William and Fazel, Amin and Holleman, Jeremy and Huang, Xinyuan and Hurtado, Robert and Kanter, David and Lokhmotov, Anton and Patterson, David and Pau, Danilo and Seo, Jae-sun and Sieracki, Jeff and Thakker, Urmish and Verhelst, Marian and Warden, Pete},
  booktitle={Advances in Neural Information Processing Systems (NeurIPS), Datasets and Benchmarks Track},
  volume={34},
  year={2021},
  publisher={Curran Associates, Inc.},
  arxiv={2106.07597},
  url={https://proceedings.neurips.cc/paper/2021/hash/fd3528311751f7a03804598927488574-Abstract.html}
}

@article{david2021tensorflow,
  title={TensorFlow Lite Micro: Embedded Machine Learning on TinyML Systems},
  author={David, Robert and Duke, Jared and Jain, Advait and Reddi, Vijay Janapa and Jeffries, Nat and Li, Jian and Kreeger, Nick and Nappier, Ian and Natraj, Meghna and Wang, Tiezhen and Warden, Pete and Rhodes, Rocky},
  journal={Proceedings of Machine Learning and Systems (MLSys)},
  volume={3},
  pages={800--811},
  year={2021},
  url={https://proceedings.mlsys.org/paper/2021/hash/9fe77ac706edb76e4e13183fa9f3b1c8-Abstract.html},
  arxiv={2010.08678}
}

@article{ray2022tinyml,
  title={A Review on TinyML: State-of-the-art and Prospects},
  author={Ray, Partha Pratim},
  journal={Journal of King Saud University - Computer and Information Sciences},
  volume={34},
  number={4},
  pages={1595--1623},
  year={2022},
  publisher={Elsevier},
  doi={10.1016/j.jksuci.2021.11.019}
}

@article{moody2001impact,
  title={The Impact of the MIT-BIH Arrhythmia Database},
  author={Moody, George B. and Mark, Roger G.},
  journal={IEEE Engineering in Medicine and Biology Magazine},
  volume={20},
  number={3},
  pages={45--50},
  year={2001},
  publisher={IEEE},
  doi={10.1109/51.932724},
  url={https://ieeexplore.ieee.org/document/932724}
}

@article{goldberger2000physiobank,
  title={PhysioBank, PhysioToolkit, and PhysioNet: Components of a New Research Resource for Complex Physiologic Signals},
  author={Goldberger, A. and Amaral, L. and Glass, L. and Hausdorff, J. and Ivanov, P. C. and Mark, R. and Mietus, J. E. and Moody, G. B. and Peng, C. K. and Stanley, H. E.},
  journal={Circulation},
  volume={101},
  number={23},
  pages={e215--e220},
  year={2000},
  publisher={American Heart Association},
  doi={10.1161/01.CIR.101.23.e215}
}

@article{hannun2019cardiologist,
  title={Cardiologist-Level Arrhythmia Detection and Classification in Ambulatory Electrocardiograms Using a Deep Neural Network},
  author={Hannun, Awni Y. and Rajpurkar, Pranav and Haghpanahi, Masouneh and Tison, Geoffrey H. and Bourn, Codie and Turakhia, Mintu P. and Ng, Andrew Y.},
  journal={Nature Medicine},
  volume={25},
  pages={65--69},
  year={2019},
  publisher={Nature Publishing Group},
  doi={10.1038/s41591-018-0268-3}
}

@article{ribeiro2020automatic,
  title={Automatic Diagnosis of the 12-Lead ECG Using a Deep Neural Network},
  author={Ribeiro, Antonio H. and Ribeiro, Manoel Horta and Paixao, Gabriela M. M. and Oliveira, Derick M. and Gomes, Paulo R. and Canazart, Jessica A. and Lima, Milton P. S. and Pastor, Carlos A. and Andersson, Carl R. and Macfarlane, Peter W. and Meira, Wagner Jr. and Schon, Thomas B. and Ribeiro, Antonio Luiz P.},
  journal={Nature Communications},
  volume={11},
  pages={1760},
  year={2020},
  publisher={Nature Publishing Group},
  doi={10.1038/s41467-020-15432-4}
}

@article{rieke2020future,
  title={The Future of Digital Health with Federated Learning},
  author={Rieke, Nicola and Hancox, Jonny and Li, Wenqi and Milletari, Fausto and Roth, Holger R. and Albarqouni, Shadi and Bakas, Spyridon and Galtier, Mathieu N. and Landman, Bennett A. and Maier-Hein, Klaus H. and Ourselin, Sebastien and Sheller, Micah J. and Summers, Ronald M. and Trask, Andrew and Xu, Daguang and Baust, Maximilian and Cardoso, Manuel Jorge},
  journal={NPJ Digital Medicine},
  volume={3},
  pages={119},
  year={2020},
  publisher={Nature Publishing Group},
  doi={10.1038/s41746-020-00323-1}
}

@article{xu2021federated,
  title={Federated Learning for Healthcare Informatics},
  author={Xu, Jie and Glicksberg, Benjamin S. and Su, Chang and Walker, Peter and Bian, Jiang and Wang, Fei},
  journal={Journal of Healthcare Informatics Research},
  volume={5},
  number={1},
  pages={1--19},
  year={2021},
  publisher={Springer},
  doi={10.1007/s41666-020-00082-4}
}

@article{brisimi2018federated,
  title={Federated Learning of Predictive Models from Federated Electronic Health Records},
  author={Brisimi, Theodora S. and Chen, Ruidi and Mela, Theofanie and Olshevsky, Alex and Paschalidis, Ioannis Ch. and Shi, Wei},
  journal={International Journal of Medical Informatics},
  volume={112},
  pages={59--67},
  year={2018},
  publisher={Elsevier},
  doi={10.1016/j.ijmedinf.2018.01.007}
}

@inproceedings{lin2017focal,
  title={Focal Loss for Dense Object Detection},
  author={Lin, Tsung-Yi and Goyal, Priya and Girshick, Ross and He, Kaiming and Doll{\'a}r, Piotr},
  booktitle={Proceedings of the IEEE International Conference on Computer Vision (ICCV)},
  pages={2980--2988},
  year={2017},
  publisher={IEEE},
  doi={10.1109/ICCV.2017.324}
}

@inproceedings{zhu2019deep,
  title={Deep Leakage from Gradients},
  author={Zhu, Ligeng and Liu, Zhijian and Han, Song},
  booktitle={Advances in Neural Information Processing Systems (NeurIPS)},
  volume={32},
  year={2019},
  url={https://proceedings.neurips.cc/paper/2019/hash/60a6c4002cc7b29142def8871531281a-Abstract.html}
}
